\documentclass[letterpaper]{article} % DO NOT CHANGE THIS
\usepackage{aaai2026}  % DO NOT CHANGE THIS
\usepackage{times}  % DO NOT CHANGE THIS
\usepackage{helvet}  % DO NOT CHANGE THIS
\usepackage{courier}  % DO NOT CHANGE THIS
\usepackage[hyphens]{url}  % DO NOT CHANGE THIS
\usepackage{graphicx} % DO NOT CHANGE THIS
\usepackage{natbib}  % DO NOT CHANGE THIS AND DO NOT ADD ANY OPTIONS TO IT
\usepackage{caption} % DO NOT CHANGE THIS AND DO NOT ADD ANY OPTIONS TO IT
\usepackage{algorithm}
\usepackage{algorithmic}

\usepackage{newfloat}
\usepackage{listings}

\usepackage{multirow}
\usepackage{booktabs}

\DeclareCaptionStyle{ruled}{labelfont=normalfont,labelsep=colon,strut=off} % DO NOT CHANGE THIS
\floatstyle{ruled}
\newfloat{listing}{tb}{lst}{}
\floatname{listing}{Listing}
\title{AncientBench: Towards Comprehensive Evaluation on Excavated and Transmitted Chinese Corpora}
\author{Zhihan Zhou$^{1,2}$, Daqian Shi$^{2,3}$, Rui Song$^{1,2}$, Lida Shi$^{2,4}$, Xiaolei Diao\thanks{Corresponding authors}$^{1,2,5}$, Hao Xu\footnotemark[1]$^{1,2}$}
\affiliations{$^1$College of Computer Science and Technology, Jilin University \\
$^2$Key Laboratory of Ancient Chinese Script, Culture Relics and Artificial Intelligence, Jilin University \\
$^3$Digital Environment Research Institute, Queen Mary University of London \\
$^4$School of Artificial Intelligence, Jilin University \\
$^5$Department of Information Engineering and Computer Science, University of Trento \\
\{zhzhou25, shild21\}@mails.jlu.edu.cn, d.shi@qmul.ac.uk, \{songrui, xuhao\}@jlu.edu.cn, xiaolei.diao@unitn.it \\}

\usepackage{bibentry}
\begin{document}

\maketitle

\begin{abstract}
Comprehension of ancient texts plays an important role in archaeology and understanding of Chinese history and civilization. The rapid development of large language models needs benchmarks that can evaluate their comprehension of ancient characters. Existing Chinese benchmarks are mostly targeted at modern Chinese and transmitted documents in ancient Chinese, but the part of excavated documents in ancient Chinese is not covered. To meet this need, we propose the AncientBench, which aims to evaluate the comprehension of ancient characters, especially in the scenario of excavated documents. The AncientBench is divided into four dimensions, which correspond to the four competencies of ancient character comprehension: glyph comprehension, pronunciation comprehension, meaning comprehension, and contextual comprehension. The benchmark also contains ten tasks, including radical, phonetic radical, homophone, cloze, translation, and more, providing a comprehensive framework for evaluation. We convened archaeological researchers to conduct experimental evaluations, proposed an ancient model as baseline, and conducted extensive experiments on the currently best-performing large language models. The experimental results reveal the great potential of large language models in ancient textual scenarios as well as the gap with humans. Our research aims to promote the development and application of large language models in the field of archaeology and ancient Chinese language.
\end{abstract}
% 我们召集考古学研究人员进行了实验评估，{同时提出了一个Ancient模型作为baseline}，并在目前性能表现最好的大语言模型上进行了广泛的实验，实验结果揭示了大语言模型在古文场景下的巨大潜力以及与人类的差距，旨在促进大语言模型在考古和古汉语领域的发展和应用。


\section{Introduction}

As one of the most spoken languages in the world, Chinese has received extensive attention in recent years in the field of natural language processing (NLP). As an important part of the Chinese language, ancient Chinese carries extremely rich historical and cultural information, and its study is of great significance for the traceability of Chinese history, the protection of cultural heritage, and the development of historical linguistics. However, traditional ancient Chinese research methods are highly dependent on the researcher's memory and linguistic intuition, which is often inefficient and difficult to deal with large-scale corpus systematically, limiting the breadth and depth of research.

\begin{figure}[!t]
    \centering
    \includegraphics[width=0.47\textwidth]{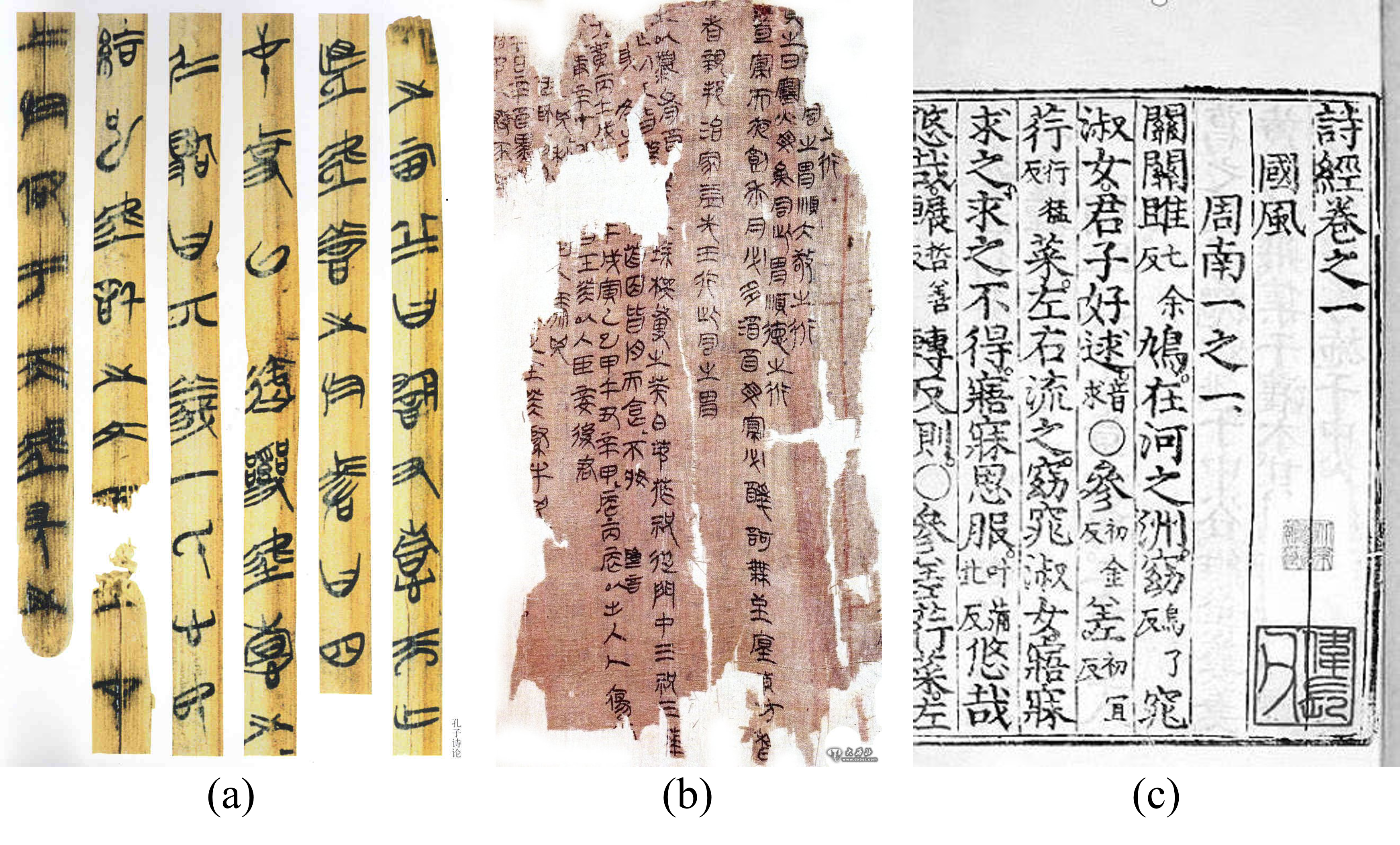}
    \caption{Comparison of excavated documents with transmitted documents. (a) Bamboo Book of Chu, excavated documents. (b) Silk Books, excavated documents. (c) Book of Poetry, transmitted documents.}
    \label{fig:Excavated Documents}
\end{figure}

With the rapid development of artificial intelligence technology, especially large language models (LLMs), ancient Chinese research is gradually stepping towards a new paradigm of data-driven and model-supported. LLMs have demonstrated excellent generalization ability in natural language understanding and natural language generation, and have shown great potential in text analysis, language structure modeling, and so on. In order to systematically evaluate the capability of these models, academics have gradually constructed a series of standardized evaluation systems, such as MMLU\cite{2021Measuring}, BIG-bench\cite{Srivastava2022BeyondTI}, and HELM\cite{2023Holistic}, etc., which have become important tools for measuring the general language intelligence of models. This technological trend offers new opportunities for ancient Chinese processing, especially in understanding difficult ancient texts.
% This technological trend brings unprecedented opportunities for the automated processing of ancient Chinese, especially for difficult ancient text comprehension tasks.

In the Chinese context, based on its linguistic characteristics and application requirements, a series of benchmarks have emerged to evaluate the capability of Chinese LLMs, such as CLUE\cite{xu-etal-2020-clue}, CMMLU\cite{li-etal-2024-cmmlu}, and MMCU\cite{Zeng_2023}, etc. Most of these benchmarks cover a wide range of domains in Chinese scenarios, such as language comprehension, logical reasoning, instruction execution, etc., which promote the evaluation and iterative optimization of Chinese language models. In recent years, some datasets related to ancient Chinese have also appeared, e.g., ACLUE\cite{zhang-li-2023-large} and WYWEB\cite{zhou-etal-2023-wyweb}. However, the tasks related to ancient Chinese in such benchmarks are scarce, mostly limited to syntax parsing, with no unified evaluation criteria, making it hard to fully assess the comprehension ability of language models for ancient Chinese. 
% However, the tasks related to ancient Chinese in these benchmark are very limited, and most of them are sentence structure parsing tasks, and the evaluation criteria of these datasets have not yet been unified, which makes it difficult to comprehensively reflect the comprehension ability of language models for ancient Chinese.

Ancient Chinese documents are mainly divided into two categories: transmitted documents (TraDoc) and excavated documents (ExcDoc) \cite{ExcavatedDocuments}. Transmitted documents refer to those that have been handed down from ancient times through hand-copying, engraving, copying, etc., such as the Analects, the Grand Scribe's Records, and the Book of Poetry. These documents have been continuously collated and annotated by future generations, forming a relatively stable textual system with standardized language and clear structure, which facilitates the construction of standardized comprehension and reasoning tasks, and is the main source of corpus for most of the current ancient literature evaluation tasks. In contrast, excavated documents refer to ancient written materials obtained through archaeological excavations, mainly including oracle bones, Bronze Inscriptions, Bamboo Book of Chu, Silk Books, and so on. 
% These documents are much earlier in time and have a more primitive and authentic language style, which can more directly reflect the state of language use and cultural practices in ancient societies. 
These documents are much earlier in time and have a more primitive and authentic language style, directly reflecting language use and culture in ancient societies. 
We give a detailed comparison between transmitted documents and excavated documents in Figure \ref{fig:Excavated Documents}.

On the one hand, excavated documents are mostly obtained from original and unearthed objects \cite{chi2022zinet} that have not been copied, compiled, or politically interfered with over the generations, thus preserving the most primitive form and content of the text \cite{boltz1986early}. On the other hand, excavated documents usually have clear archaeological layers, coexisting artifacts, and scientific dating data, which can accurately locate their age \cite{jane2014ancient}. Therefore, excavated documents have stronger originality, authenticity, and clarity of time and region than transmitted documents. Excavated documents are of irreplaceable importance in the study of ancient Chinese for the understanding of ancient documents. Existing LLMs in the field of ancient Chinese still mainly rely on transmitted documents in the training and evaluation process, and the coverage of excavated documents is extremely limited because most of the texts of excavated documents are concentrated on the surface of tortoise shells and bamboo slips, which are difficult to access and covered with complex noises \cite{shi2022rcrn}.
% \cite{diao2025oracle}. 
Some attempts include the use of OCR technology to detect and recognize excavated documents \cite{diao2023rzcr, yue2025ancient}, but only a small amount of data is available because the existing text encoding in computers cannot cover all excavated documents. The above difficulties have hindered the research of LLMs in the field of ancient Chinese.

To bridge this gap, we propose AncientBench, a benchmark for evaluating the comprehension of ancient texts, especially excavated documents, in large language models. From the perspective of data corpus, AncientBench mainly adopts the pre-Qin period, i.e., the Xia dynasty, Shang dynasty, Zhou dynasty, Spring and Autumn periods, and Warring States period, focusing mainly on Oracle Bone Inscriptions, Bronze Inscriptions, and Bamboo Book of Chu. From the perspective of the competencies evaluated and the design of the questions, the AncientBench examines the four competencies of LLMs in the field of ancient Chinese, i.e., glyph comprehension, pronunciation comprehension, meaning comprehension, and contextual comprehension. The AncientBench contains ten tasks, i.e., Radical, Radical Meaning, Pronunciation, Phonetic Radical, Homophone, ExcDoc Word, TraDoc Word, Cloze, Phonetic Loan Character, and Translation. We aim to comprehensively and objectively evaluate the ability of LLMs to understand ancient texts.

% \begin{table*}[]
% \centering
% \resizebox{1\textwidth}{!}{
%     \begin{tabular}{cccc}
%     \hline
%     \textbf{Competencies} & \textbf{Task} & \textbf{Questions} & \textbf{Description}  \\  \hline
%     \multirow{2}{*}{glyph} &  Radical & 8438 & Radical Recognition  \\ 
%                    & Radical Meaning  & 1432 & Radical Meaning Recognition   \\ \hline 
%     \multirow{3}{*}{Pronunciation} &  Pronunciation & 3886 & Pronunciation Recognition  \\ 
%                    & Phonetic Radical  & 304 & Phonetic Radical Recognition   \\ 
%                    & Homophone  & 2265 & Homophone Recognition   \\ \hline 
%     \multirow{2}{*}{Meaning} & ExcDoc Word & 365 & Excavated Documents Word Understanding  \\ 
%                    & TraDoc Word  & 1504 & transmitted Documents Word Understanding   \\ \hline 
%     \multirow{3}{*}{Contextual} &  Cloze & 4875 & Cloze Task for Excavated Documents \\ 
%                    & Phonetic Loan Character  & 4637 & Phonetic Loan Character Recognition   \\ 
%                    & Translation  & 1001 & Translation   \\ \hline 
%         \end{tabular}
%         }
%         \caption{Framework of AncientBench. \textbf{Competencies} denotes the four competencies examined by the benchmark, e.g. glyph denotes glyph comprehension, \textbf{Task} denotes the ten tasks, \textbf{Questions} denotes the number of questions included in each task, and \textbf{Description} denotes the description of the task.}
%     \label{table:benchmark framework}
% \end{table*}

\begin{table*}[]
\centering
\resizebox{0.95\textwidth}{!}{
    \begin{tabular}{cccc}
    \hline
    \textbf{Competencies} & \textbf{Task} & \textbf{Questions} & \textbf{Description}  \\  \hline
    \multirow{2}{*}{Glyph} &  Radical & 8438 & Radical Recognition  \\ 
                   & Radical Meaning  & 1432 & Radical Meaning Recognition   \\ \hline 
    \multirow{3}{*}{Pronunciation} &  Pronunciation & 3886 & Pronunciation Recognition  \\ 
                   & Phonetic Radical  & 304 & Phonetic Radical Recognition   \\ 
                   & Homophone  & 2265 & Homophone Recognition   \\ \hline 
    \multirow{2}{*}{Meaning} & ExcDoc Word & 365 & Excavated Documents Word Understanding  \\ 
                   & TraDoc Word  & 1504 & Transmitted Documents Word Understanding   \\ \hline 
    \multirow{3}{*}{Contextual} &  Cloze & 4875 & Cloze Task for Excavated Documents \\ 
                   & Phonetic Loan Character  & 4637 & Phonetic Loan Character Recognition   \\ 
                   & Translation  & 1001 & Translation   \\ \hline 
        \end{tabular}
        }
        \caption{Framework of AncientBench. \textbf{Competencies} denotes the four competencies examined by the benchmark, e.g. Glyph denotes glyph comprehension. \textbf{Task} denotes the ten tasks. \textbf{Questions} denotes the number of questions included in each task. \textbf{Description} denotes the description of the task.}
    \label{table:benchmark framework}
\end{table*}

We convened researchers from the interdisciplinary fields of archaeology and artificial intelligence to conduct experiments based on the AncientBench to evaluate the comprehension of ancient characters as humans, and evaluated LLMs with zero-shot and few-shot, respectively. The experimental results show that there is still a gap between their comprehension of ancient texts and that of humans, and most of the models still need to improve their comprehension of ancient texts. Our study aims to propose a comprehensively and multi-dimensional benchmark in the field of ancient Chinese, with the aim of promoting the research of LLMs in the field of ancient Chinese. Our contribution is summarized below.

\begin{itemize}
\item We have achieved the digitization of ancient characters. We propose a three-stage method for digitization. With this method, we can process excavated documents, which facilitates the training and evaluation of ancient LLMs.
\item We construct AncientBench. For the first time, we introduce excavated documents into the field of natural language processing, construct AncientBench focusing on excavated documents, and propose a novel and comprehensive evaluation benchmark for the research of LLMs in the field of ancient Chinese.
\item We convened archaeological researchers to evaluate the comprehension of ancient characters in humans. At the same time, we proposed an ancient model and conducted extensive experiments based on AncientBench.
\end{itemize}
% 我们基于 AnientBench 评估了人类和 LLM 对古文字的理解能力。
% 基于该数据集，我们召集了考古学领域研究人员进行了人类古文理解能力的评估，同时我们提出了一个ancient模型，并在目前性能表现最好的大语言模型上进行了广泛的实验。

\section{Related Work}

% \subsection{Chinese benchmarks for LLMs}
% The development of benchmarks for evaluating the language comprehension ability of models has been an important research effort in the field of NLP. Early models mainly relied on pre-training fine-tuning methods and were mostly for NLU, in order to evaluate the ability of such models in NLP, a series of benchmarks have been proposed for NLU tasks, such as SentEval\cite{conneau-kiela-2018-senteval}, GLUE\cite{wang-etal-2018-glue} and SuperGLUE\cite{NEURIPS2019}, where CLUE\cite{CLUE} contains more than 10 tasks covering most NLP problems. With the development of LLs in recent years, more benchmarks focusing on NLG tasks have been proposed. For example Sakaguchi \cite{10.1145/3474381}, Hendrycks\cite{Hendrycks}, Austin\cite{Austin}. MMLU is extremely popular in LLM evaluations because of its standardised question format, comprehensive coverage and uniformity of evaluation criteria. However, most of the benchmarks mentioned above focus on evaluating in English.
The development of benchmarks for evaluating models has been an important research topic in NLP. Early models mainly relied on pretraining fine-tuning methods, and a series of benchmarks have been proposed for NLU tasks, such as SentEval\cite{conneau-kiela-2018-senteval}, GLUE\cite{wang-etal-2018-glue} and SuperGLUE\cite{NEURIPS2019_4496bf24}, where CLUE\cite{CLUE} contains more than 10 tasks covering most NLP problems. With the development of LLMs in recent years, A series of benchmarks have been proposed for NLG tasks, such as Sakaguchi\cite{10.1145/3474381}, Hendrycks\cite{Hendrycks}, and Austin\cite{Austin}. MMLU\cite{2021Measuring} is extremely popular in LLM evaluations due to its standardized question format, comprehensive coverage, and uniformity of the evaluation criteria. However, most of the benchmarks mentioned above focus on evaluating in English.

As the most spoken language in the world, Chinese has a very important research significance. Xu proposed CLUE\cite{CLUE} and SuperCLUE\cite{xu2023supercluecomprehensivechineselarge}, which cover most of the tasks of Chinese NLU. After MMLU\cite{2021Measuring} was proposed, many researchers proposed Chinese scenarios based on the MMLU evaluation framework. Multi-domain and multidimensional benchmarks, such as MMCU\cite{Zeng_2023}, CMMLU\cite{li-etal-2024-cmmlu}, and AGIEval\cite{zhong-etal-2024-agieval}, where MMCU covers four major domains and CMMLU covers more than 20 multilingual and Chinese languages. In addition, CG-Eval\cite{zeng2024evaluatinggenerationcapabilitieslarge} and M3KE\cite{liu2023m3kemassivemultilevelmultisubject} used multitask multiple question types to evaluate LLMs.

In recent years, benchmarks for evaluating ancient Chinese have made rich progress in several directions. \cite{zinin-xu-2020-corpus} further open-sourced 20 historical traveling materials and other ancient corpus, enriching the diversity of the domain; FSPC\cite{FSPC} and CcMP\cite{li2021ccpmchineseclassicalpoetry} focus on the comprehension task of ancient poems, and the CUGE\cite{yao2022cugechineselanguageunderstanding} focuses on the poetry matching subtask based on CcMP. In addition, \cite{pan-etal-2022-zuo}, \cite{wang-ren-2022-uncertainty}, and \cite{Liu} have successively introduced task sets covering syntactic analysis, topic mining, and sentiment classification; Comprehensive benchmarks such as CCLUE\cite{yao2022cugechineselanguageunderstanding} and WYWEB\cite{zhou-etal-2023-wyweb} integrate various tasks from text classification to poetry analysis and machine reading comprehension. AC-EVAL\cite{wei-etal-2024-ac} is mainly targeted at LLMs, and integrates a variety of tasks and datasets, constructing a comprehensive and integrated benchmark for ancient Chinese.

\section{AncientBench}
\subsection{AncientBench Overview}
AncientBench consists of four dimensions of competency evaluations and ten tasks, as shown in Table \ref{table:benchmark framework}. The four dimensions of competency evaluation include glyph comprehension, pronunciation comprehension, meaning comprehension, and contextual comprehension. The ten tasks include Radical, Radical Meaning, Pronunciation, Phonetic Radical, Homophone, ExcDoc Word, TraDoc Word, Cloze, Phonetic Loan Character, and Translation. In order to ensure the quality of the competency dimensions and task definitions, we used the following criteria for the definitions.

The first is authority and credibility. We invited experts in the field of archaeology to analyze the comprehension of ancient characters from a human perspective, i.e., what are the foundational abilities that persons need if we consider them to have the ability to understand ancient texts. Combining archaeologists' analysis and classical tasks in the field of natural language processing, we finally identified these four dimensions.

The second is diversity and comprehensiveness. In order to ensure the comprehensiveness of the AncientBench, we defined it from two perspectives: the characteristics of ancient Chinese documents and the difficulty of textual comprehension. The four dimensions of the AncientBench represent a surface-to-surface understanding of an ancient Chinese character, where glyph understanding and pronunciation understanding represent the external features of a character. 
The glyph is key to transferring the semantics of characters \cite{shi2022charformer}. 
In glyph comprehension, Radical Recognition refers to the recognition of the components of a character \cite{diao2023toward}, and Radical Meaning refers to the recognition towards the meaning of the components. In pronunciation comprehension, Pronunciation Recognition refers to the recognition of the pronunciation of a character, and we refer to the notation in the ancient Chinese dictionary ShuoWen\cite{shuowenjiezi} as the definition of character pronunciation, while Phonetic Radical Recognition refers to the recognition of the component of a character that is responsible for the pronunciation of the character \cite{ho2003radical}. Meaning comprehension and contextual comprehension are the internal meanings of a character \cite{nelson2007fostering}. In meaning comprehension, ExcDoc Word refers to the understanding of the word meanings in excavated documents, and TraDoc Word refers to the understanding of the word meanings in transmitted documents. Contextual comprehension involves identifying Phonetic Loan Characters \cite{yang2000} in excavated texts. AncientBench's four tasks vary in difficulty: glyph and pronunciation comprehension is at the character level, meaning comprehension is at the word level, and contextual comprehension is at the sentence level.
% In contextual comprehension, Phonetic Loan Character refers to the recognition of phonetic loan character in excavated documents. The four dimensions of the AncientBench are also differentiated in terms of difficulty from a natural language processing perspective. Glyph comprehension and pronunciation comprehension are at the character level, meaning comprehension is at the word level, and contextual comprehension is at the sentence level.

Finally, the standardization of the criteria. In order to ensure the uniformity and certainty of the evaluation criteria of AncientBench. We chose the classic tasks in the field of natural language processing, such as cloze, translation, etc. as the evaluation tasks. Then we further processed these tasks and finally displayed them in multiple-choice format to ensure the consistency and certainty of the evaluation criteria. AncientBench contains 28,707 questions, all in multiple-choice format.

\subsection{AncientBench Construction}
\subsubsection{Data Collection.}
The data sources for the AncientBench mainly include oracle bone inscription, Bronze Inscriptions, Bamboo Book of Chu, ShuoWen\cite{shuowenjiezi}, Chinese Dictionary Compendium\cite{HanyuDaCidian1986}, and a series of pre-Qin books.

We have digitized excavated documents such as oracle bone inscriptions and Bamboo Book of Chu, as well as transmitted documents such as ShuoWen and Chinese Dictionary Compendium based on the constructed character coding list. Then we matched and integrated each character with its radicals, pronunciation, and character meaning information.

% \begin{figure*}[t]
%     \centering
%     \includegraphics[width=1\textwidth]{/img/image3.png}
%     \caption{Some samples from the AncientBench. The red box shows a Radical Meaning task sample for glyph comprehension, the blue box shows a Phonetic Radical task sample for pronunciation comprehension, the yellow box shows an ExcDoc Word task sample for meaning comprehension, and the green box shows a Phonetic Loan Character task sample for contextual comprehension.}
%     \label{fig:question example}
% \end{figure*}

\begin{figure*}[t]
    \centering
    \includegraphics[width=0.96\textwidth]{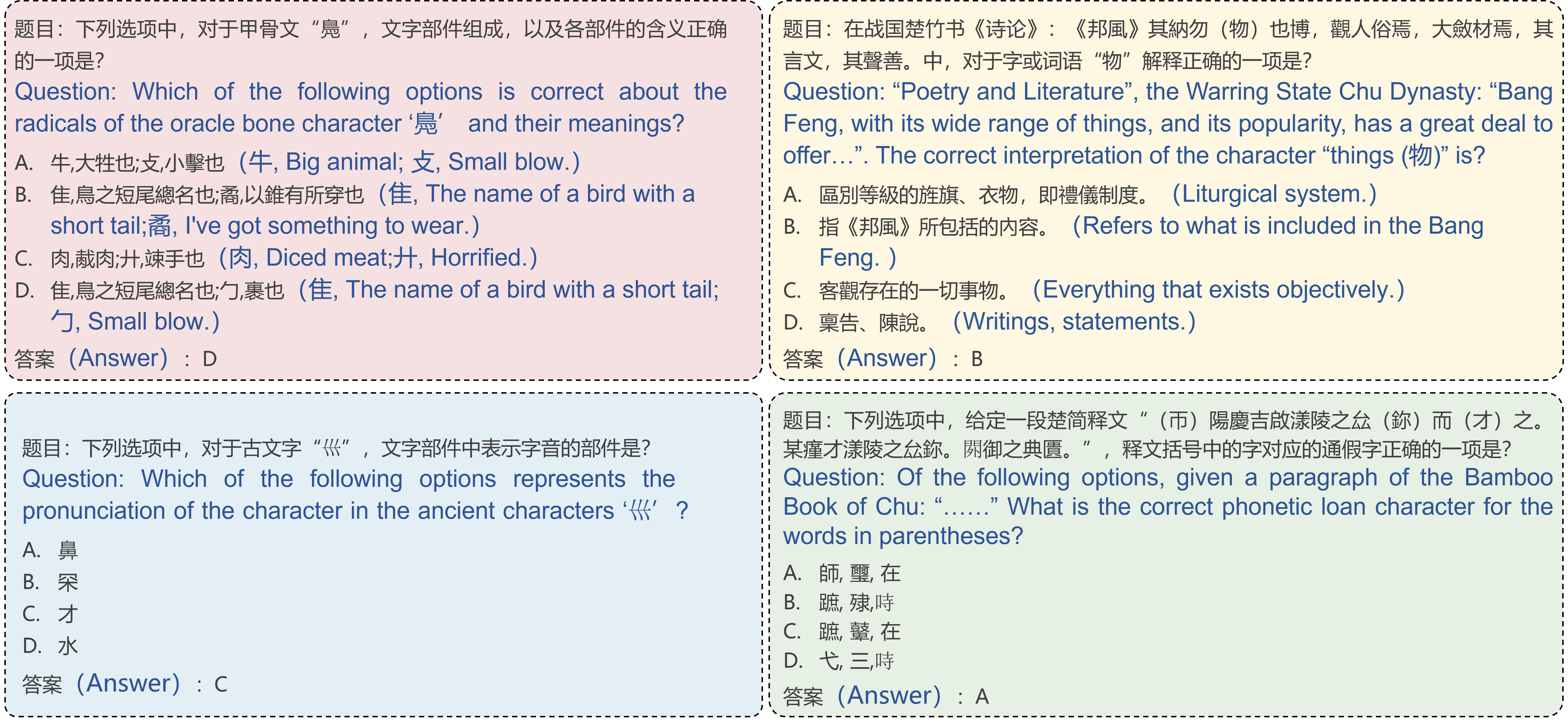}
    \caption{Some samples from the AncientBench. The red box shows a Radical Meaning task sample for glyph comprehension, the blue box shows a Phonetic Radical task sample for pronunciation comprehension, the yellow box shows an ExcDoc Word task sample for meaning comprehension, and the green box shows a Phonetic Loan Character task sample for contextual comprehension.}
    \label{fig:question example}
\end{figure*}

\subsubsection{Digitization of Ancient Characters.}
One of the main difficulties in constructing datasets of ancient characters is that many ancient characters are not encoded in computers, and many ancient characters that can be represented in computers may have different unicode encoding. In order to standardize the evaluation of ancient characters for comprehension, digitization and encoding of ancient characters need to be standardized. In this paper, we use a three-phase approach to achieve digitization and encoding of ancient characters, i.e., ancient character image processing, unified font encoding, and new encoding of missing characters.

The first is ancient character image processing. We extract the feature information and spatial relationship information of each radical of the ancient character image by computer vision methods, and link the ancient characters with the radical information to generate the ancient character knowledge graph. Then we reconstruct a vector image for each ancient character.

The second is unified font encoding. We extract glyphs and encoding information from the font library based by character processing technology, and perform de-duplication operations on characters to ensure that each character corresponds to a unique encoding information, and finally obtain an integrated character encoding table.

Finally the new encoding of missing characters. The glyph information obtained from the image processing of ancient characters is compared with the character encoding table, if it is an existing character, the unicode encoding in the character encoding table continues to be used, and if it is an unrecorded character, it is added to the vocabulary as a new character generating a new unicode encoding.

After the three-stage approach to digitization of ancient characters, we can obtain a complete and unified table for character encoding.

\subsubsection{Subject Construction.}
Based on the four dimensions we defined for capability evaluation, we analyzed the existing available data resources, as well as referring to the classic tasks in the field of natural language reasoning, and finally identified ten evaluation tasks, i.e., Radical, Radical Meaning, Pronunciation, Phonetic Radical, Homophone, ExcDoc Word, TraDoc Word, Cloze, Phonetic Loan Character, and Translation. In order to ensure the accuracy and reasonableness of the topic construction, we followed the following principles in reconstructing the subject and options. Some sample questions are shown in Figure \ref{fig:question example}.

The first is uniform evaluation criteria. In order to standardize the input and output of the model, as well as ensuring uniformity in the criteria for evaluating the model, we used multiple-choice questions for all of them.

The second is difficulty level differentiation. We differentiated the difficulty levels of the AncientBench questions from three perspectives: task type, data source, and question option design. In terms of task type, ten evaluation tasks of AncientBench cover ancient characters from the understanding of various features of a single character to the semantic understanding in the context, so the Radical is defined as a relatively easy task, while the Cloze will be defined as a more difficult task. In terms of data sources, the AncientBench's data covers multiple eras such as oracle bones, Bronze Inscriptions, and Warring States scripts, etc. Scripts from the ancient period will have greater differences in the way of using words and in the glyph and pronunciation of words compared to those of more recent eras, so even though the ExcDoc Word belongs to the meaning comprehension ability, and the Translation belongs to the contextual comprehension ability, ExcDoc Word will be more difficult than Translation, because the words in ExcDoc Word is older than that in Translation. In the same task, different subjects also have difficulty differentiation, for example, the Radical includes Oracle Bone, Bronze Inscriptions, and Warring States scripts. Oracle Bone is more difficult to identify compared to Warring States script because the glyphs of Oracle Bone are quite different from modern Chinese characters. As for the design of the question options, in order to avoid the question options being too intuitive, we did not generate all the question options by random method, but carefully designed the difficulty differentiation. For some of the options, we replaced a radical of the correct answer with a similar radical to ensure the reasonableness of the questions, as shown in the red box in Figure \ref{fig:question example}, when we constructed option B, we chose the part of the answer that has a similar meaning to the correct answer as the confusing option. We further refine the distinction between models or human ancient characters by differentiating difficulty between tasks, and within tasks.

Finally the authority. After the AncientBench was constructed, we invited researchers in the field of archaeology to evaluate the AncientBench from the aspects of data source selection, topic design, and option design to ensure the reasonableness of the AncientBench.

\section{Experiments}

\subsection{Setup}
We evaluated both the LLMs and human comprehension of ancient characters based on AncientBench, and for the LLMs, we compared the accuracy of the models in both zero-shot and few-shot, respectively.

Referring to the experimental setup of CMMLU\cite{li-etal-2024-cmmlu}, AC-EVAL\cite{wei-etal-2024-ac}, we used the following statement in constructing the model input: ``For the following multiple-choice questions, please directly give the option of the correct answer". For zero-shot, we will directly give the output prompt: ``the answer is: ". For few-shot, we will give five samples containing the correct answer.

Regarding the output of the model, we refer to the answer processing method of MMLU\cite{2021Measuring}, the logits of the next of the next predicted token is obtained after inputting the prompt into the model, then we compare the probability of the four tokens, `A', `B', `C' and ‘D’, and finally we choose the one with the highest probability as the model choice. 

For the matrices, we calculated multiple-choice accuracy for the ten tasks, averaged the accuracy for the tasks included in the four competencies as the score for each competency, and finally calculated the average of all competency scores as the average score.

\begin{table*}[]
\centering
\resizebox{0.90\textwidth}{!}{
    \begin{tabular}{cccccc}
    \toprule
    \textbf{Model} & \textbf{Glyph} & \textbf{Pronunciation} & \textbf{Meaning } & \textbf{Contextual} & \textbf{Average} \\ \midrule
        human performance & 76.66 & 50.00 & 38.33 & 55.55 & 55.13 \\
        Qwen-14B-Chat\cite{bai2023qwentechnicalreport} & \textbf{53.01} & 31.70 & 63.86 & 55.43 & \textbf{51.00} \\ 
        GLM4-9b-chat\cite{glm2024chatglmfamilylargelanguage} & 44.96 & 34.48 & \textbf{69.71} & 48.05 & 49.30 \\
        Baichuan2-13B-Chat\cite{yang2025baichuan2openlargescale} & 45.08 & \textbf{35.16} & 60.81 & 56.11 & 49.29 \\
        Yi1.5-9B-Chat\cite{ai2025yiopenfoundationmodels} & 42.70 & 30.12 & 65.53 & 56.91 & 48.81 \\
        Baichuan2-7B-Chat\cite{yang2025baichuan2openlargescale} & 49.32 & 29.68 & 58.54 & 54.64 & 48.04 \\
        Llama3-8B-Instruct\cite{grattafiori2024llama3herdmodels} & 41.54 & 28.15 & 51.80 & \textbf{62.41} & 45.97 \\
        Xunzi-Qwen-Chat\cite{XunziALLM} & 43.03 & 27.40 & 55.23 & 49.72 & 43.84 \\
        Qwen-7B-Chat\cite{bai2023qwentechnicalreport} & 45.80 & 25.08 & 56.33 & 48.14 & 43.83 \\ 
        Tonggu-7b-chat\cite{cao2024tonggumasteringclassicalchinese} & 33.72 & 29.37 & 42.49 & 38.65 & 36.05 \\ \midrule
        Yi1.5-9B-Ancient(ours) & 50.77 & 32.30 & 67.12 & 50.93 & 50.28 \\ \bottomrule
        \end{tabular}
        }
        \caption{Accuracy of each model in zero-shot. We report the average of each tasks in comprehension capacity. ``Average" is the average of the comprehension competencies. Bold indicates the best model performance.}
    \label{table:zero-shot}
\end{table*}

\subsubsection{Human Performance.}
To facilitate comparison with the model, we evaluated the comprehension of ancient characters in humans. In conducting the evaluation, we randomly selected five multiple-choice questions for each of the 10 tasks, constituting a 50-question multiple-choice questionnaire, and then invited the participants to take part in the examination. We counted the participants' accuracy and time to answer the questions, and took the average as the average of the accuracy of each task. The invited participants were graduate students at the intersection of the fields of archaeology and artificial intelligence.

\subsubsection{Models.}

In our evaluation, we have selected 9 models that perform well in the Chinese field. Including (1) Llama-3-8B-Instruct\cite{grattafiori2024llama3herdmodels}, (2) GLM-4-9b-Chat\cite{glm2024chatglmfamilylargelanguage}, (3) Qwen-7B/14B-Chat\cite{bai2023qwentechnicalreport}, (4) Baichuan2-7B/13B-Chat\cite{yang2025baichuan2openlargescale}, (5) Yi-1.5-9B-Chat\cite{ai2025yiopenfoundationmodels}, (6) Xunzi-Qwen-Chat\cite{XunziALLM}, (7) Tonggu-7b-Chat\cite{cao2024tonggumasteringclassicalchinese}.

% our model
% 为了评估古文知识对大语言模型古文理解能力的影响，同时为benchmark设立一个强大的baseline，我们基于古文知识对大语言模型进行了微调。
% 根据我们的前期实验，我们使用了中文能力最强的模型，即Yi1.5-9B-Chat。我们构建了古文能力指令微调数据集，具体来说，。基于该指令微调数据集，我们对大语言模型进行了全量微调。
\subsection{Ancient Model}
In order to evaluate the impact of ancient Chinese knowledge on the ancient Chinese comprehension ability of LLMs, and to establish a strong baseline for benchmark, we fine-tuned LLMs based on ancient Chinese knowledge.

Based on our preliminary experiments, we used Yi1.5-9B-Chat\cite{ai2025yiopenfoundationmodels}, which has the strongest ancient Chinese capabilities. We constructed a fine-tuning dataset. Specifically, we constructed fine-tuning question-answer pairs, each consisting of three parts: instruction, input, and output. The instruction describes what task to perform, the input describes the specific task details, and the output defines the model output. Based on this dataset, we performed full fine-tuning on Yi1.5-9B-Chat. During training, we set the batch size to 2 and the learning rate to $1e - 5$. All experiments were executed on the machine running the Ubuntu OS with ascend-d910b npu. We named our model Yi1.5-9B-Ancient.

% 基于这个微调数据集，我们对 Yi1.5-9B-Chat 进行了全面微调。在训练过程中，我们将batch_size设置为2，学习率设置为 1e - 5。我们还使用 PyTorch 框架制作实验模型，所有实验均在运行 Ubuntu 操作系统的计算机系统上执行，该系统配备4个ascend-d910b npu。我们将我们的模型命名为 Yi1.5-9B-Ancient。

\subsection{Zero-shot Results}
% We evaluated the comprehension of ancient characters by LLMs in zero-shot and few-shot, and compared with human performance. We analysed the experimental results in terms of the number of parameters of the model, the type of model and the different tasks.
We evaluated the comprehension of ancient characters by LLMs in zero-shot, and compared with human performance. We analyzed the results in terms of the number of parameters of the model, the type of model, and the different tasks.

% \subsubsection{Zero-shot}
The ancient characters comprehension of the LLMs in zero-shot is shown in Table \ref{table:zero-shot}. The average competence of ancient characters comprehension of LLMs is lower than the human performance, and even the best performing model, i.e., Qwen-14B-Chat, is 4.13\% lower than the average human level. For glyph comprehension and pronunciation comprehension, there is a large difference between the LLMs and the human performance, and the human performance is generally more than 30\% higher than the LLMs, which may be because humans are multimodal in recognizing ancient characters' glyphs, whereas the LLMs have only a single modality. For tasks such as context comprehension, which requires only a single modality, there is not much difference between the LLMs and the human performance. For meaning comprehension, the LLMs generally scored higher than the human performance, probably because the word sense comprehension task requires more memory.

From the perspective of model type, the average ancient characters comprehension competence of the ancient Chinese model and the generic LLMs did not differ much. In terms of the average score, Qwen-14B-Chat, Baichuan2-13B-Chat, and GLM4-9b-chat have the highest scores of 51.00\%, 49.30\%, and 49.29\% respectively, which shows that although most of these models are trained on a modern Chinese corpus, they possess strong generalization ability. Llama3-8B-Instruct's is trained mainly on the English corpus, and thus scores slightly lower than the Chinese model with the same number of parameters, which is in line with our conjecture. With the same number of parameters, the glyph comprehension and pronunciation comprehension of the ancient Chinese model is slightly higher than that of the generic LLM, which may be because of the larger number of ancient characters included in the training corpus of the ancient Chinese model, which enables the model to build up information between the ancient characters and their glyph. The above observations reflect the linguistic differences between English and Chinese as well as between modern and ancient Chinese in the training corpora, further emphasizing the importance of our benchmarks.

From the perspective of the number of model parameters, the highest average score is Qwen-14B-Chat with 51.00\%, which shows that increasing the number of model parameters is beneficial to the comprehension of ancient characters, the average score of Baichuan2-13B-Chat is 1.25\% higher than that of Baichuan2-7B-Chat, and Qwen-14B-Chat's average score improved by 7.17\% over Qwen-7B-Chat. For meaning comprehension and contextual comprehension, the effect of the number of parameters is more obvious, with Baichuan2-13B-Chat's meaning comprehension being 2.27\% higher than Baichuan2-7B-Chat's, and Qwen-14B-Chat's contextual comprehension being 7.29\% higher than Qwen-7B-Chat's, which may be because meaning comprehension and contextual comprehension require more logical reasoning skills compared to Glyph and Pronunciation comprehension.

\begin{table*}[]
\centering
\resizebox{0.95\textwidth}{!}{
    \begin{tabular}{cccccc}
    \toprule
    \textbf{Model} & \textbf{Glyph} & \textbf{Pronunciation} & \textbf{Meaning } & \textbf{Contextual} & \textbf{Average} \\ \midrule
        human performance & 76.66 & 50.00 & 38.33 & 55.55 & 55.13 \\
        Qwen-14B-Chat\cite{bai2023qwentechnicalreport} & 47.31 (-5.7) & 29.75 (-1.95) & 65.08 (+1.22) & 58.10 (+2.67) & 50.05 (-0.95) \\ 
        GLM4-9b-chat\cite{glm2024chatglmfamilylargelanguage} & 46.33 (+1.37) & 34.92 (+0.44) & \textbf{69.84} (+0.13) & 51.05 (+3.00) & 50.53 (+1.23) \\
        Baichuan2-13B-Chat\cite{yang2025baichuan2openlargescale} & 43.84 (-1.24) & \textbf{35.34} (+0.18) & 57.13 (-3.68) & 55.37 (-0.74) & 47.92 (-1.37) \\
        Yi1.5-9B-Chat\cite{ai2025yiopenfoundationmodels} & 46.73 (+4.03) & 32.04 (+1.92) & 66.21 (+0.68) & 58.26 (+1.35) & 50.80 (+1.99) \\
        Baichuan2-7B-Chat\cite{yang2025baichuan2openlargescale} & 42.74 (-6.58) & 32.10 (+2.42) & 56.78 (-1.76) & 50.54 (-4.1) & 45.54 (-2.5) \\
        Llama3-8B-Instruct\cite{grattafiori2024llama3herdmodels} & 41.24 (-0.3) & 27.10 (-1.14) & 57.17 (+5.37) & \textbf{64.57} (+2.16) & 47.51 (+1.54) \\
        Xunzi-Qwen-Chat\cite{XunziALLM} & 44.63 (+1.6) & 29.20 (+1.8) & 52.67 (-2.56) & 49.03 (-0.69) & 43.88 (+0.04) \\
        Qwen-7B-Chat\cite{bai2023qwentechnicalreport} & 42.53 (-3.72) & 25.77 (+0.69) & 55.47 (-0.86) & 49.37 (+1.23) & 43.28 (-0.55) \\ 
        Tonggu-7b-chat\cite{cao2024tonggumasteringclassicalchinese} & 34.54 (+0.82) & 30.41 (+1.04) & 45.34 (+2.85) & 39.09 (+0.44) & 37.34 (+1.29) \\  \midrule
        Yi1.5-9B-Ancient(ours) & \textbf{49.67} (-1.1) & 32.98 (+0.68) & 65.23 (-1.89) & 55.44 (+4.51) & \textbf{50.83} (+0.55) \\ \bottomrule
        \end{tabular}
        }
        \caption{Accuracy of each model in few-shot. We report the average of each tasks in comprehension capacity. ``Average" is the average of the comprehension competencies. Bold indicates the best model performance.}
    \label{table:few-shot}
\end{table*}

From the perspective of AncientBench's different tasks, the LLMs scored high in the evaluation of meaning comprehension and contextual comprehension competence, with GLM4-9b-chat reaching 69.71\% in meaning comprehension, which is 31.38\% higher than the human performance; however, they performed poorly in the evaluation of glyph comprehension and pronunciation comprehension ability, with all the models scoring around 30 in the pronunciation comprehension evaluations, close to random selection. This may be because the questions give the context, and the models are able to use this information to reason the meanings of the ancient characters and then provide answers; whereas glyph comprehension and pronunciation comprehension directly require the models to recognize the glyph and pronunciation, which relies more on the amount of knowledge inherent in the models, i.e., the knowledge learned during the pre-training process. The above conclusion proves that the reasoning ability and the amount of knowledge of LLMs can indeed assist us in the research in the field of ancient characters, but the current LLMs have less knowledge related to excavated documents, which is caused by the low content of excavated documents in the pre-training corpus of the models, which shows that the LLMs have a great potential in the research in the field of ancient Chinese, and also side by side proves the importance of our proposed benchmarks.

It is worth noting that Yi1.5-9B-Ancient, after being fine-tuned with our ancient Chinese knowledge, demonstrated excellent performance. The average score reached 50.28\%, achieving the best performance among LLMs with the same parameter scale, and an average score increase of 1.47\% compared to the Yi1.5-9B-Chat. In terms of glyph comprehension, Yi1.5-9B-Ancient achieved 50.77\%, an improvement of 8.07\% over Yi1.5-9B-Chat. In terms of pronunciation comprehension and meaning comprehension, Yi1.5-9B-Ancient achieved 32.30\% and 67.12\%, respectively. The above results prove that Yi1.5-9B-Chat contains relatively little information on ancient characters, and it is likely that LLMs such as Yi1.5-9B-Chat have limited embedding of ancient characters, while the method of fine-tuning can supplement this part of the knowledge of large language models. Yi1.5-9B-Ancient performed poorly on the contextual comprehension task, even 5.98\% lower than Yi1.5-9B-Chat. This maybe because the contextual comprehension contains some topics related to traditional documents, and the ancient knowledge introduced by fine-tuning caused some damage to the original embedding of Yi1.5-9B-Chat, thereby affecting the model's performance on the contextual comprehension. In summary, existing LLMs lack knowledge related to ancient Chinese, resulting in weak ancient character comprehension abilities. Fine-tuning as a baseline is an intuitive and effective method that can supplement LLMs with ancient Chinese knowledge, but it may affect the original embedding. Therefore, it is necessary to further propose better methods that can represent ancient characters without affecting the original embedding.

% 值得注意的是Yi1.5-9B-Ancient经过我们古文知识微调，展现出了较好的性能。平均得分达到了50.28%，在相同参数规模的大语言模型中达到了最好的性能，并且相较于Yi1.5-9B-Chat模型平均得分提升了1.47%。在字形理解方面Yi1.5-9B-Ancient达到了50.77%，比Yi1.5-9B-Chat提升了8.07%，并且对于字音理解和Meaning，Yi1.5-9B-Ancient分别达到了32.30%和67.12%。以上结果证明了Yi1.5-9B-Chat中包含的古文字字形信息较少，很可能Yi1.5-9B-Chat等大语言模型的古文字的embedding有限，而指令微调的方法可以一定程度上补充大语言模型这部分知识。在Contextual任务上Yi1.5-9B-Ancient表现一般，甚至比Yi1.5-9B-Chat降低了5.98%，这可能是由于Contextual任务Contextual任务中包含了一些传世文献相关的题目，而指令微调引入的古文字相关的知识对Yi1.5-9B-Chat原有的embedding造成了一定的破坏，从而影响了模型在上下文任务上的表现。综上所述，现有的大语言模型缺乏古文领域相关的知识，使得古文字理解能力较弱。指令微调作为baseline是一个直观有效的方法可以一定程度上为大语言模型补充古文相关知识，但是可能会影响原有的embedding。
% 因此有必要进一步提出更好的可以表征古文字且不影响原有embedding的方法。 

\subsection{Few-shot Results}
% \subsubsection{Few-shot}
We analyzed the few-shot results in terms of the number of parameters of the model, the type of model, and the different tasks, and compared the results with the zero-shot results.

The ancient characters comprehension of the LLMs in few-shot is shown in Table \ref{table:few-shot}. Compared to zero-shot, the performance of most of the models is improved, but still lower than the human performance. Yi1.5-9B-Chat achieves a average score improvement of 1.99\% to 50.80\%. tonggu-7b-chat, glm-4-9b-chat, and yi-1.5-9B-Chat in few-shot have all the four  competence scores all improved. However, both Baichuan2 and Qwen have reduced performance in few-shot, which is different from our expectation.

From the point of view of each task, the model's pronunciation comprehension and contextual comprehension improved in few-shot, and the model's pronunciation comprehension score in zero-shot was close to 25\%, i.e., randomly selected, which may be because the model was unfamiliar with the content of the questions and the options, and the samples given in the model inputs in the few-shot experiments helped the model's comprehension, which led to the model's improved performance.

The Yi1.5-9B-Ancient model achieved 50.83\% in few-shot, which is the best performance among all LLMs, an improvement of 0.03\% over the Yi1.5-9B-Chat, and a 2.94\% improvement in the glyph comprehension task score.

% Yi1.5-9B-Ancient模型在few-shot的情况下达到了50.83%，是所有大语言模型中最好的性能，比微调前Yi1.5-9B-Chat提升了0.03%，字形理解任务得分提升了2.94%%。

\section{Conclusion}
In this paper, we present AncientBench, a benchmark centered on excavated documents containing four competencies and ten tasks that comprehensively evaluates the comprehension of ancient characters for large language models. We propose a three-stage approach to digitize ancient characters and encode them into computers. In addition, we propose an ancient model as a baseline and evaluate human performance and large language models based on AncientBench. Our research introduces excavated documents into the field of natural language processing for the first time.

In future work, we will further expand the sources of data and extend it to multiple modalities. In addition, we will explore more effective model evaluation methods and metrics. Our research will provide a good foundation for large language models in the field of ancient Chinese.

% 本文中我们提出了AncientBench，一个以出土文献为中心的benchmark，包含4项能力和十项任务，能够全面的评估大语言模型的古文字理解能力。我们提出了一个三阶段的古文字数字化的方法，将古文字编码到计算机中。此外，我们提出了一个ancient model作为baseline，并基于AncientBench对人类表现和大语言模型进行了评估和分析。我们的研究首次将出土文献引入自然语言处理领域。

% 未来的工作中，我们会对数据的来源进一步扩充，并扩展到多种模态。此外，我们还会探索更有效的模型评估方法和指标。我们的研究会为大语言模型在古文领域的一眼就提供一个良好的基础。

\section{Acknowledgments}
This research is supported by the National Natural Science Foundation of China (No.62476111), the Department of Science and Technology of Jilin Province, China (20230201086GX), the ``Paleography and Chinese Civilization Inheritance and Development Program" Collaborative Innovation Platform (No.G3829), the National Social Science Foundation of China (No. 23VRC033), and the interdisciplinary cultivation project for young teachers and students at Jilin University, China (No. 2024-JCXK-04).

\bibliography{aaai2026}

@inproceedings{2021Measuring,
  title={Measuring Massive Multitask Language Understanding},
  author={ Hendrycks, Dan  and  Burns, Collin  and  Basart, Steven  and  Zou, Andy  and  Mazeika, Mantas  and  Song, Dawn  and  Steinhardt, Jacob },
  booktitle={International Conference on Learning Representations},
  year={2021},
}

@article{Srivastava2022BeyondTI,
  title={Beyond the Imitation Game: Quantifying and extrapolating the capabilities of language models},
  author={Srivastava, Aarohi and Rastogi, Abhinav and Rao, Abhishek and Shoeb, Abu Awal Md and Abid, Abubakar and Fisch, Adam and Brown, Adam R and Santoro, Adam and Gupta, Aditya and Garriga-Alonso, Adri{\`a} and others},
  journal={TRANSACTIONS ON MACHINE LEARNING RESEARCH},
  year={2022}
}

@article{2023Holistic,
  title={Holistic Evaluation of Language Models},
  author={ Bommasani, Rishi  and  Liang, Percy  and  Lee, Tony },
  journal={Annals of the New York Academy of Sciences},
  volume={1525},
  number={1},
  year={2023},
}

@inproceedings{xu-etal-2020-clue,
    title = "{CLUE}: A {C}hinese Language Understanding Evaluation Benchmark",
    author = "Xu, Liang  and
      Hu, Hai  and
      Zhang, Xuanwei  and
      Li, Lu  and
      Cao, Chenjie  and
      Li, Yudong  and
      Xu, Yechen  and
      Sun, Kai  and
      Yu, Dian  and
      Yu, Cong  and
      Tian, Yin  and
      Dong, Qianqian  and
      Liu, Weitang  and
      Shi, Bo",
    booktitle = "Proceedings of the 28th International Conference on Computational Linguistics",
    year = "2020",
    doi = "10.18653/v1/2020.coling-main.419",
    pages = "4762--4772",
}

@inproceedings{li-etal-2024-cmmlu,
    title = "{CMMLU}: Measuring massive multitask language understanding in {C}hinese",
    author = "Li, Haonan  and
      Zhang, Yixuan  and
      Koto, Fajri  and
      Yang, Yifei  and
      Zhao, Hai  and
      Gong, Yeyun  and
      Duan, Nan  and
      Baldwin, Timothy",
    booktitle = "Findings of the Association for Computational Linguistics: ACL 2024",
    year = "2024",
    doi = "10.18653/v1/2024.findings-acl.671",
    pages = "11260--11285",
}

@article{Zeng_2023,  
 title={Measuring Massive Multitask Chinese Understanding}, 
 author={Zeng, Hui}, 
 year={2023}, 
 month={Apr}, 
 journal={ArXiv},
 language={en-US} 
 }

@inproceedings{zhang-li-2023-large,
    title = "Can Large Language Model Comprehend {A}ncient {C}hinese? A Preliminary Test on {ACLUE}",
    author = "Zhang, Yixuan  and
      Li, Haonan",
    editor = "Anderson, Adam  and
      Gordin, Shai  and
      Li, Bin  and
      Liu, Yudong  and
      Passarotti, Marco C.",
    booktitle = "Proceedings of the Ancient Language Processing Workshop",
    month = sep,
    year = "2023",
    address = "Varna, Bulgaria",
    publisher = "INCOMA Ltd., Shoumen, Bulgaria",
    url = "https://aclanthology.org/2023.alp-1.9/",
    pages = "80--87",
}

@inproceedings{zhou-etal-2023-wyweb,
    title = "{WYWEB}: A {NLP} Evaluation Benchmark For Classical {C}hinese",
    author = "Zhou, Bo  and
      Chen, Qianglong  and
      Wang, Tianyu  and
      Zhong, Xiaomi  and
      Zhang, Yin",
    editor = "Rogers, Anna  and
      Boyd-Graber, Jordan  and
      Okazaki, Naoaki",
    booktitle = "Findings of the Association for Computational Linguistics: ACL 2023",
    month = jul,
    year = "2023",
    address = "Toronto, Canada",
    publisher = "Association for Computational Linguistics",
    url = "https://aclanthology.org/2023.findings-acl.204/",
    doi = "10.18653/v1/2023.findings-acl.204",
    pages = "3294--3319",
}

@inproceedings{conneau-kiela-2018-senteval,
    title = "{S}ent{E}val: An Evaluation Toolkit for Universal Sentence Representations",
    author = "Conneau, Alexis  and
      Kiela, Douwe",
    editor = "Calzolari, Nicoletta  and
      Choukri, Khalid  and
      Cieri, Christopher  and
      Declerck, Thierry  and
      Goggi, Sara  and
      Hasida, Koiti  and
      Isahara, Hitoshi",
    booktitle = "Proceedings of the Eleventh International Conference on Language Resources and Evaluation ({LREC} 2018)",
    month = may,
    year = "2018",
    address = "Miyazaki, Japan",
    publisher = "European Language Resources Association (ELRA)",
    url = "https://aclanthology.org/L18-1269/"
}

@inproceedings{wang-etal-2018-glue,
    title = "{GLUE}: A Multi-Task Benchmark and Analysis Platform for Natural Language Understanding",
    author = "Wang, Alex  and
      Singh, Amanpreet  and
      Michael, Julian  and
      Hill, Felix  and
      Levy, Omer  and
      Bowman, Samuel",
    editor = "Linzen, Tal  and
      Chrupa{\l}a, Grzegorz  and
      Alishahi, Afra",
    booktitle = "Proceedings of the 2018 {EMNLP} Workshop {B}lackbox{NLP}: Analyzing and Interpreting Neural Networks for {NLP}",
    month = nov,
    year = "2018",
    address = "Brussels, Belgium",
    publisher = "Association for Computational Linguistics",
    url = "https://aclanthology.org/W18-5446/",
    doi = "10.18653/v1/W18-5446",
    pages = "353--355",
}

@inproceedings{NEURIPS2019_4496bf24,
 author = {Wang, Alex and Pruksachatkun, Yada and Nangia, Nikita and Singh, Amanpreet and Michael, Julian and Hill, Felix and Levy, Omer and Bowman, Samuel},
 booktitle = {Advances in Neural Information Processing Systems},
 editor = {H. Wallach and H. Larochelle and A. Beygelzimer and F. d\textquotesingle Alch\'{e}-Buc and E. Fox and R. Garnett},
 pages = {},
 publisher = {Curran Associates, Inc.},
 title = {SuperGLUE: A Stickier Benchmark for General-Purpose Language Understanding Systems},
 url = {https://proceedings.neurips.cc/paper_files/paper/2019/file/4496bf24afe7fab6f046bf4923da8de6-Paper.pdf},
 volume = {32},
 year = {2019}
}

@inproceedings{CLUE,
author = {Xu, Liang and Hu, Hai and Zhang, Xuanwei and Li, Lu and Cao, Chenjie and Li, Yudong and Xu, Yechen and Sun, Kai and Yu, Dian and Yu, Cong and Tian, Yin and Dong, Qianqian and Liu, Weitang and Shi, Bo and Cui, Yiming and Li, Junyi and Zeng, Jun and Wang, Rongzhao and Xie, Weijian and Lan, Zhenzhong},
year = {2020},
month = {01},
pages = {4762-4772},
title = {CLUE: A Chinese Language Understanding Evaluation Benchmark},
doi = {10.18653/v1/2020.coling-main.419},
}

@article{10.1145/3474381,
author = {Sakaguchi, Keisuke and Bras, Ronan Le and Bhagavatula, Chandra and Choi, Yejin},
title = {WinoGrande: an adversarial winograd schema challenge at scale},
year = {2021},
issue_date = {September 2021},
publisher = {Association for Computing Machinery},
address = {New York, NY, USA},
volume = {64},
number = {9},
issn = {0001-0782},
url = {https://doi.org/10.1145/3474381},
doi = {10.1145/3474381},
journal = {Commun. ACM},
month = aug,
pages = {99–106},
numpages = {8}
}

@inproceedings{Hendrycks ,
 author = {Hendrycks, Dan and Burns, Collin and Kadavath, Saurav and Arora, Akul and Basart, Steven and Tang, Eric and Song, Dawn and Steinhardt, Jacob},
 booktitle = {Proceedings of the Neural Information Processing Systems Track on Datasets and Benchmarks},
 editor = {J. Vanschoren and S. Yeung},
 pages = {},
 title = {Measuring Mathematical Problem Solving With the MATH Dataset},
 url = {https://datasets-benchmarks-proceedings.neurips.cc/paper_files/paper/2021/file/be83ab3ecd0db773eb2dc1b0a17836a1-Paper-round2.pdf},
 volume = {1},
 year = {2021}
}

@article{Austin,
  author = {Jacob Austin and Augustus Odena and
  Maxwell I. Nye and Maarten Bosma and
  Henryk Michalewski and David Dohan and
  Ellen Jiang and Carrie J. Cai and
  Michael Terry and Quoc V. Le and Charles Sutton},
  title = {Program Synthesis with Large Language Models},
  journal = {CoRR},
  volume   = {abs/2108.07732},
  year = {2021},
  url = {https://arxiv.org/abs/2108.07732},
  eprinttype  = {arXiv},
  eprint   = {2108.07732},
  bibsource  = {dblp computer science bibliography, https://dblp.org}
}

@misc{xu2023supercluecomprehensivechineselarge,
      title={SuperCLUE: A Comprehensive Chinese Large Language Model Benchmark}, 
      author={Liang Xu and Anqi Li and Lei Zhu and Hang Xue and Changtai Zhu and Kangkang Zhao and Haonan He and Xuanwei Zhang and Qiyue Kang and Zhenzhong Lan},
      year={2023},
      eprint={2307.15020},
      archivePrefix={arXiv},
      primaryClass={cs.CL},
      url={https://arxiv.org/abs/2307.15020}, 
}

@inproceedings{zhong-etal-2024-agieval,
    title = "{AGIE}val: A Human-Centric Benchmark for Evaluating Foundation Models",
    author = "Zhong, Wanjun  and
      Cui, Ruixiang  and
      Guo, Yiduo  and
      Liang, Yaobo  and
      Lu, Shuai  and
      Wang, Yanlin  and
      Saied, Amin  and
      Chen, Weizhu  and
      Duan, Nan",
    editor = "Duh, Kevin  and
      Gomez, Helena  and
      Bethard, Steven",
    booktitle = "Findings of the Association for Computational Linguistics: NAACL 2024",
    month = jun,
    year = "2024",
    address = "Mexico City, Mexico",
    publisher = "Association for Computational Linguistics",
    url = "https://aclanthology.org/2024.findings-naacl.149/",
    doi = "10.18653/v1/2024.findings-naacl.149",
    pages = "2299--2314",
}

@misc{zeng2024evaluatinggenerationcapabilitieslarge,
      title={Evaluating the Generation Capabilities of Large Chinese Language Models}, 
      author={Hui Zeng and Jingyuan Xue and Meng Hao and Chen Sun and Bin Ning and Na Zhang},
      year={2024},
      eprint={2308.04823},
      archivePrefix={arXiv},
      primaryClass={cs.CL},
      url={https://arxiv.org/abs/2308.04823}, 
}

@misc{liu2023m3kemassivemultilevelmultisubject,
      title={M3KE: A Massive Multi-Level Multi-Subject Knowledge Evaluation Benchmark for Chinese Large Language Models}, 
      author={Chuang Liu and Renren Jin and Yuqi Ren and Linhao Yu and Tianyu Dong and Xiaohan Peng and Shuting Zhang and Jianxiang Peng and Peiyi Zhang},
      year={2023},
      eprint={2305.10263},
      archivePrefix={arXiv},
      primaryClass={cs.CL},
      url={https://arxiv.org/abs/2305.10263}, 
}

@inproceedings{zinin-xu-2020-corpus,
    title = "Corpus of {C}hinese Dynastic Histories: Gender Analysis over Two Millennia",
    author = "Zinin, Sergey  and
      Xu, Yang",
    editor = "Calzolari, Nicoletta  and
      B{\'e}chet, Fr{\'e}d{\'e}ric  and
      Blache, Philippe  and
      Choukri, Khalid  and
      Cieri, Christopher  and
      Declerck, Thierry  and
      Goggi, Sara  and
      Isahara, Hitoshi  and
      Maegaard, Bente  and
      Mariani, Joseph  and
      Mazo, H{\'e}l{\`e}ne  and
      Moreno, Asuncion  and
      Odijk, Jan  and
      Piperidis, Stelios",
    booktitle = "Proceedings of the Twelfth Language Resources and Evaluation Conference",
    month = may,
    year = "2020",
    address = "Marseille, France",
    publisher = "European Language Resources Association",
    url = "https://aclanthology.org/2020.lrec-1.98/",
    pages = "785--793",
    language = "eng",
    ISBN = "979-10-95546-34-4",
}

@inproceedings{FSPC,
author = {Shao, Yizhan and Shao, Tong and Wang, Minghao and Wang, Peng and Gao, Jie},
title = {A Sentiment and Style Controllable Approach for Chinese Poetry Generation},
year = {2021},
isbn = {9781450384469},
publisher = {Association for Computing Machinery},
address = {New York, NY, USA},
url = {https://doi.org/10.1145/3459637.3481964},
doi = {10.1145/3459637.3481964},
booktitle = {Proceedings of the 30th ACM International Conference on Information \& Knowledge Management},
pages = {4784–4788},
numpages = {5},
location = {Virtual Event, Queensland, Australia},
series = {CIKM '21}
}

@misc{li2021ccpmchineseclassicalpoetry,
      title={CCPM: A Chinese Classical Poetry Matching Dataset}, 
      author={Wenhao Li and Fanchao Qi and Maosong Sun and Xiaoyuan Yi and Jiarui Zhang},
      year={2021},
      eprint={2106.01979},
      archivePrefix={arXiv},
      primaryClass={cs.CL},
      url={https://arxiv.org/abs/2106.01979}, 
}

@misc{yao2022cugechineselanguageunderstanding,
      title={CUGE: A Chinese Language Understanding and Generation Evaluation Benchmark}, 
      author={Yuan Yao and Qingxiu Dong and Jian Guan and Boxi Cao and Zhengyan Zhang and Chaojun Xiao and Xiaozhi Wang and Fanchao Qi and Junwei Bao and Jinran Nie and Zheni Zeng and Yuxian Gu and Kun Zhou and Xuancheng Huan},
      year={2022},
      eprint={2112.13610},
      archivePrefix={arXiv},
      primaryClass={cs.CL},
      url={https://arxiv.org/abs/2112.13610}, 
}

@inproceedings{pan-etal-2022-zuo,
    title = "Zuo Zhuan {A}ncient {C}hinese Dataset for Word Sense Disambiguation",
    author = "Pan, Xiaomeng  and
      Wang, Hongfei  and
      Oka, Teruaki  and
      Komachi, Mamoru",
    editor = "Ippolito, Daphne  and
      Li, Liunian Harold  and
      Pacheco, Maria Leonor  and
      Chen, Danqi  and
      Xue, Nianwen",
    booktitle = "Proceedings of the 2022 Conference of the North American Chapter of the Association for Computational Linguistics: Human Language Technologies: Student Research Workshop",
    month = jul,
    year = "2022",
    address = "Hybrid: Seattle, Washington + Online",
    publisher = "Association for Computational Linguistics",
    url = "https://aclanthology.org/2022.naacl-srw.17/",
    doi = "10.18653/v1/2022.naacl-srw.17",
    pages = "129--135"
}

@inproceedings{wang-ren-2022-uncertainty,
    title = "The Uncertainty-based Retrieval Framework for {A}ncient {C}hinese {CWS} and {POS}",
    author = "Wang, Pengyu  and
      Ren, Zhichen",
    editor = "Sprugnoli, Rachele  and
      Passarotti, Marco",
    booktitle = "Proceedings of the Second Workshop on Language Technologies for Historical and Ancient Languages",
    month = jun,
    year = "2022",
    address = "Marseille, France",
    publisher = "European Language Resources Association",
    url = "https://aclanthology.org/2022.lt4hala-1.25/",
    pages = "164--168"
}

@article{Liu,
author = {Liu, Maofu and Xiang, Junyi and Xia, Xu and Hu, Huijun},
title = {Contrastive Learning between Classical and Modern Chinese for Classical Chinese Machine Reading Comprehension},
year = {2022},
issue_date = {February 2023},
publisher = {Association for Computing Machinery},
address = {New York, NY, USA},
volume = {22},
number = {2},
issn = {2375-4699},
url = {https://doi.org/10.1145/3551637},
doi = {10.1145/3551637},
journal = {ACM Trans. Asian Low-Resour. Lang. Inf. Process.},
month = dec,
articleno = {49},
numpages = {22}
}

@inproceedings{wei-etal-2024-ac,
    title = "{AC}-{EVAL}: Evaluating {A}ncient {C}hinese Language Understanding in Large Language Models",
    author = "Wei, Yuting  and
      Xu, Yuanxing  and
      Wei, Xinru  and
      Yangsimin, Yangsimin  and
      Zhu, Yangfu  and
      Li, Yuqing  and
      Liu, Di  and
      Wu, Bin",
    editor = "Al-Onaizan, Yaser  and
      Bansal, Mohit  and
      Chen, Yun-Nung",
    booktitle = "Findings of the Association for Computational Linguistics: EMNLP 2024",
    month = nov,
    year = "2024",
    address = "Miami, Florida, USA",
    publisher = "Association for Computational Linguistics",
    url = "https://aclanthology.org/2024.findings-emnlp.87/",
    doi = "10.18653/v1/2024.findings-emnlp.87",
    pages = "1600--1617",
}

@misc{grattafiori2024llama3herdmodels,
      title={The Llama 3 Herd of Models}, 
      author={Aaron Grattafiori and Abhimanyu Dubey and Abhinav Jauhri and Abhinav Pandey and Abhishek Kadian and Ahmad Al-Dahle and Aiesha Letman and Akhil Mathur and Alan Schelten and Alex Vaughan and Amy Yang and Angela Fan and Anirudh Goyal and Anthony Hartshorn and Aobo Yang},
      year={2024},
      eprint={2407.21783},
      archivePrefix={arXiv},
      primaryClass={cs.AI},
      url={https://arxiv.org/abs/2407.21783}, 
}

@article{glm2024chatglmfamilylargelanguage,
  title={ChatGLM: A Family of Large Language Models from GLM-130B to GLM-4 All Tools},
  author={Zeng, Aohan and Xu, Bin and Wang, Bowen and Zhang, Chenhui and Yin, Da and Rojas, Diego and Feng, Guanyu and Zhao, Hanlin and Lai, Hanyu and Yu, Hao and others},
  journal={CoRR},
  year={2024}
}

@misc{bai2023qwentechnicalreport,
      title={Qwen Technical Report}, 
      author={Jinze Bai and Shuai Bai and Yunfei Chu and Zeyu Cui and Kai Dang and Xiaodong Deng and Yang Fan and Wenbin Ge and Yu Han and Fei Huang and Binyuan Hui and Luo Ji and Mei Li and Junyang Lin and Runji Lin and Dayiheng Liu and Gao Liu and Chengqiang Lu and Keming Lu and Jianxin Ma},
      year={2023},
      eprint={2309.16609},
      archivePrefix={arXiv},
      primaryClass={cs.CL},
      url={https://arxiv.org/abs/2309.16609}, 
}

@misc{yang2025baichuan2openlargescale,
      title={Baichuan 2: Open Large-scale Language Models}, 
      author={Aiyuan Yang and Bin Xiao and Bingning Wang and Borong Zhang and Ce Bian and Chao Yin and Chenxu Lv and Da Pan and Dian Wang and Dong Yan and Fan Yang and Fei Deng and Feng Wang and Feng Liu and Guangwei Ai and Guosheng Dong and Haizhou Zhao and Hang Xu and Haoze Sun},
      year={2025},
      eprint={2309.10305},
      archivePrefix={arXiv},
      primaryClass={cs.CL},
      url={https://arxiv.org/abs/2309.10305}, 
}

@book{ExcavatedDocuments,
    author  = {Wang Guiyuan},
    title   = {A Study of Excavated Documents in China},
    year    = "2023",
    publisher = {Routledge},
    address = {London},
    doi = "https://doi.org/10.4324/9781032623078",
}

@inproceedings{chi2022zinet,
  title={ZiNet: Linking Chinese Characters Spanning Three Thousand Years},
  author={Chi, Yang and Giunchiglia, Fausto and Shi, Daqian and Diao, Xiaolei and Li, Chuntao and Xu, Hao},
  booktitle={Findings of the Association for Computational Linguistics: ACL 2022},
  pages={3061--3070},
  year={2022}
}

@misc{jane2014ancient,
  title={Ancient times table hidden in Chinese bamboo strips},
  author={Jane, Q},
  year={2014}
}

@article{boltz1986early,
  title={Early chinese writing},
  author={Boltz, William G},
  journal={World Archaeology},
  volume={17},
  number={3},
  pages={420--436},
  year={1986},
  publisher={Taylor \& Francis}
}

@inproceedings{shi2022charformer,
  title={CharFormer: A Glyph Fusion based Attentive Framework for High-precision Character Image Denoising},
  author={Shi, Daqian and Diao, Xiaolei and Shi, Lida and Tang, Hao and Chi, Yang and Li, Chuntao and Xu, Hao},
  booktitle={Proceedings of the 30th ACM International Conference on Multimedia},
  year={2022}
}

@inproceedings{shi2022rcrn,
  title={RCRN: Real-world Character Image Restoration Network via Skeleton Extraction},
  author={Shi, Daqian and Diao, Xiaolei and Tang, Hao and Li, Xiaomin and Xing, Hao and Xu, Hao},
  booktitle={Proceedings of the 30th ACM International Conference on Multimedia},
  year={2022}
}

@inproceedings{diao2023rzcr,
  title={RZCR: Zero-shot Character Recognition via Radical-based Reasoning},
  author={Diao, Xiaolei and Shi, Daqian and Tang, Hao and Shen, Qiang and Li, Yanzeng and Wu, Lei and Xu, Hao},
  year={2023},
  booktitle={IJCAI},
}

@inproceedings{diao2023toward,
  title={Toward Zero-shot Character Recognition: A Gold Standard Dataset with Radical-level Annotations},
  author={Diao, Xiaolei and Shi, Daqian and Li, Jian and Shi, Lida and Yue, Mingzhe and Qi, Ruihua and Li, Chuntao and Xu, Hao},
  booktitle={Proceedings of the 31st ACM International Conference on Multimedia},
  pages={6869--6877},
  year={2023}
}

@article{yue2025ancient,
  title={Ancient character detection based on fine-grained density map},
  author={Yue, Mingzhe and Shi, Daqian and Diao, Xiaolei and Guo, Shuzhen and Li, Chuntao and Xu, Hao},
  journal={npj Heritage Science},
  volume={13},
  number={1},
  pages={280},
  year={2025},
  publisher={Springer International Publishing Cham}
}

@article{ho2003radical,
  title={A “radical” approach to reading development in Chinese: The role of semantic radicals and phonetic radicals},
  author={Ho, Connie Suk-Han and Ng, Ting-Ting and Ng, Wing-Kin},
  journal={Journal of literacy research},
  volume={35},
  number={3},
  pages={849--878},
  year={2003},
  publisher={SAGE Publications Sage CA: Los Angeles, CA}
}

@article{nelson2007fostering,
  title={Fostering the development of vocabulary knowledge and reading comprehension though contextually-based multiple meaning vocabulary instruction},
  author={Nelson, J Ron and Stage, Scott A},
  journal={Education and treatment of children},
  volume={30},
  number={1},
  pages={1--22},
  year={2007},
  publisher={West Virginia University Press}
}

@article{yang2000,
title={A Brief Discussion on Phonetic Loan Characters},
author={Heming Yang},
journal={Journal of Wuhan University (Humanities and Social Sciences Edition)},
volume={1},
year={2000}
}

@book{shuowenjiezi,
  editor = "Shen Xu",
  title = "Shuowen Jiezi Zhu",
  year = 1981,
  address = "Shanghai",
  publisher = "Shanghai Ancient Books Publishing House",
}

@book{HanyuDaCidian1986,
  title = {Hanyu Da Cidian (The Grand Chinese Dictionary)},
  editor = {Zhufeng Luo},
  address = {Shanghai},
  publisher = {Hanyu Da Cidian Press},
  year = {1986},
  note = {Published in 12 volumes from 1986 to 1994}
}

@misc{ai2025yiopenfoundationmodels,
      title={Yi: Open Foundation Models by 01.AI}, 
      author={Alex Young and Bei Chen and Chao Li and Chengen Huang and Ge Zhang and Guanwei Zhang},
      year={2025},
      eprint={2403.04652},
      archivePrefix={arXiv},
      primaryClass={cs.CL},
      url={https://arxiv.org/abs/2403.04652}, 
}

@misc{XunziALLM,
  title = {XunziALLM},
  author = {Zhixiao Zhao and Si Shen and Bin Li and Xueliang Ma},
  url = {https://github.com/Xunzi-LLM-of-Chinese-classics/XunziALLM},
  year = 2024
}

@misc{cao2024tonggumasteringclassicalchinese,
      title={TongGu: Mastering Classical Chinese Understanding with Knowledge-Grounded Large Language Models}, 
      author={Jiahuan Cao and Dezhi Peng and Peirong Zhang and Yongxin Shi and Yang Liu and Kai Ding and Lianwen Jin},
      year={2024},
      eprint={2407.03937},
      archivePrefix={arXiv},
      primaryClass={cs.CL},
      url={https://arxiv.org/abs/2407.03937}, 
}

\end{document}